\documentclass{article}

     \PassOptionsToPackage{numbers, compress}{natbib}

\usepackage[preprint]{neurips_2024}

\usepackage[utf8]{inputenc} 
\usepackage[T1]{fontenc}    
\usepackage{hyperref}       
\usepackage{url}            
\usepackage{booktabs}       
\usepackage{amsfonts}       
\usepackage{nicefrac}       
\usepackage{microtype}      
\usepackage{xcolor}         

\usepackage{algorithm}
\usepackage{algpseudocode}

\usepackage{latexsym}
\usepackage{geometry}
\usepackage{graphicx}
\usepackage{tabularx}
\usepackage{array}
\usepackage{longtable}
\usepackage{multirow}
\usepackage{tikz}
\usepackage{wrapfig}

\title{QA-TOOLBOX: Conversational Question-Answering for process task guidance in manufacturing}

\author{%
  Ramesh Manuvinakurike \\
  \And
  Elizabeth Watkins \\
  \And
  Celal Savur\\
  \And
  Anthony Rhodes \\
  \And
  Sovan Biswas \\  
  \And
  Gesem Gudino Mejia \\
  \And
  Richard Beckwith \\
  \And
  Saurav Sahay \\
  \And
  Giuseppe Raffa \\
  \And
  Lama Nachman
}

\begin{document}

\maketitle

\begin{abstract}
  In this work we explore utilizing LLMs for data augmentation for manufacturing task guidance system. The dataset consists of representative samples of interactions with technicians working in an advanced manufacturing setting. The purpose of this work to explore the task, data augmentation for the supported tasks and evaluating the performance of the existing LLMs. We observe that that task is complex requiring understanding from procedure specification documents, actions and objects sequenced temporally. The dataset consists of 200,000+ question/answer pairs that refer to the spec document and are grounded in narrations and/or video demonstrations. We compared the performance of several popular open-sourced LLMs by developing a “baseline” using each LLM and then compared the responses in a reference-free setting using LLM-as-a-judge and compared the ratings with crowd-workers whilst validating the ratings with experts.
\end{abstract}

\section{Introduction}

\begin{wrapfigure}{r}{0.45\textwidth} 
    \centering
    \includegraphics[width=0.45\textwidth]{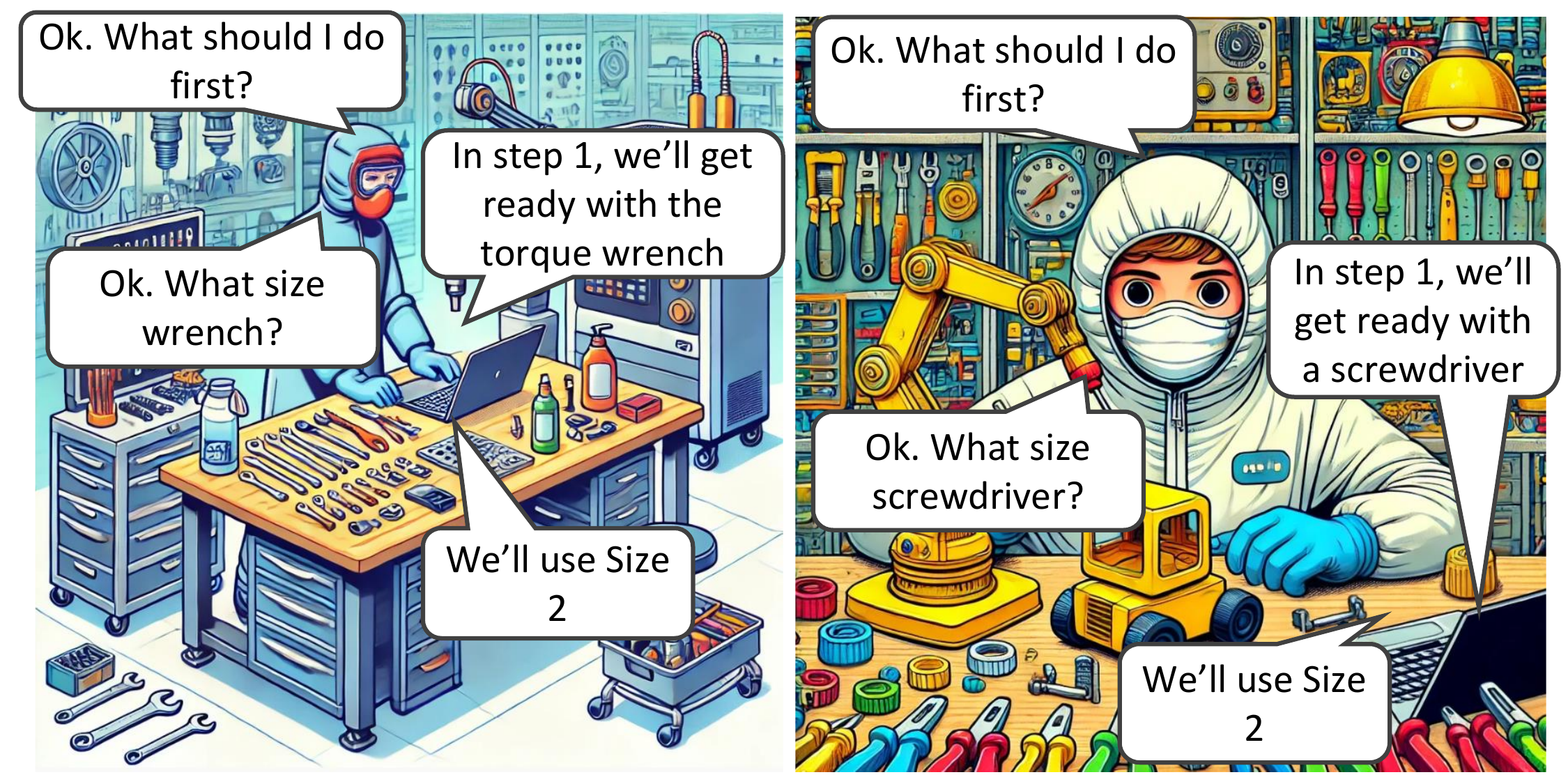}
    \caption{Example of task guidance QA exchanges.}
    \label{fig:example-label}
\end{wrapfigure}

Manufacturing is a human-centric domain where technicians must follow complex instructions provided by a \textit{spec} (task specification manual). 
A spec typically consists of natural language description of several processes that needs to be accomplished towards reaching an end goal. Manufacturing industry faces several challenges. 
Within a spec, a typical process can contain anywhere from dozens to hundreds of actions that could be carried out as-is in the specified order which can be challenging.
Furthermore, manufacturing industry has high attrition rates \cite{deloitte_manufacturing_2022}, and new workers find it challenging to perform complex tasks effectively, adding to the challenges in a successful execution of a process. 
A process task-guidance system for technicians could benefit the industry widely. Process task guidance requires the development of methods and technology for AI assistants that can help technicians perform complex  tasks \cite{darpa-ptg}. 
Task guidance with AI assistance in manufacturing remains a challenging problem  \cite{makatura2024can}.

In this work we propose a question-answering based process task guidance system, with the technician requesting information about the process and the AI assistance providing the guidance via answers. 
The process task guidance system that we are developing uses data from two main sources: \textit{specs}, from which we extract the steps described, and \textit{task executions and narrations}, which we acquire through observations as the task is undertaken. 
\textbf{Specs} include ordered instructions to be executed in order to achieve consistent outcomes. While specs are meant to be as descriptive as possible, enumerating every possible scenario is challenging. In addition, the steps are often not completely enumerated as certain actions that are local solutions  or commonsense and may not be included. 
\textbf{Task executions \& narrations} are live executions of the steps outlined in the instructions. While the task executor is expected to perform the process as mentioned in the manual, task performance can uncover novel features (e.g., the use of a different tool) or unspecified actions in the spec. Such narrations form a source of data that is valuable, if available. However, there are additional challenges involved in collecting this data.
To the best of our knowledge, a dataset that consists of question answers, demonstrations, specs and narrations do not exist. This work bridges this gap in the literature by contributing an initial version dataset (v1). We believe such a dataset could be valuable to test out the process task guidance capabilities, especially in large language (and multimodal) models. 



\begin{figure*}
    \centering
    \includegraphics[width=0.9\textwidth]{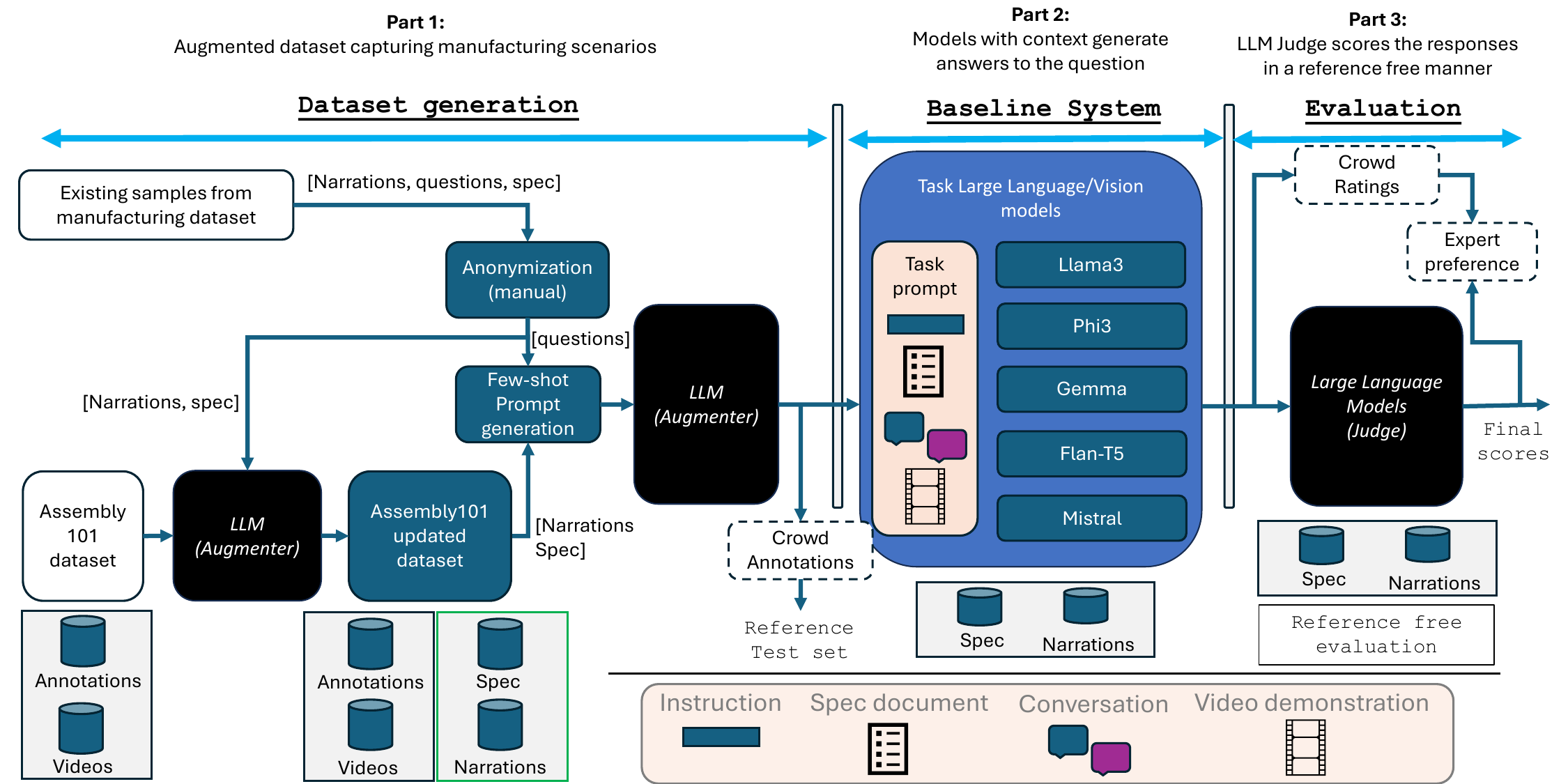}
    \caption{Workflow for creating QA-TOOLBOX dataset, baseline, and evaluations. Open-sourced dataset augmented with information needed for the manufacturing setting. Baseline systems used LLMs to generate answers. LLM-as-a-judge \cite{zheng2024judging} evaluated responses, and we performed expert validation to ensure the ratings viability.}
    \label{fig:workflow}
\end{figure*}

We also benchmark few of the available large language models encapsulating such scenarios and argue for the need of joint multimodal reasoning. 
We propose the task of Question-Answering (QA) exchanges between the technician and an AI assistant with reference to the spec document \& narrations grounded in the observed task executions. 
The task bears resemblance to document-grounded QA \cite{feng2020doc2dial}. Important distinctions here are i) the spec (document), task executions or narrations may not be all encompassing requiring a more holistic understanding of the process, ii) referring terms in the spec and speech are grounded in different demonstrations by other technicians, iii) application to manufacturing domain. 

Our work offers a novel contribution to the field by demonstrating how to develop i) a dataset and benchmark for the domain of task guidance and ii) baselines and exploring reference free evaluations using an LLM-as-a-judge \cite{zheng2024judging}. We also present a data augmentation pipeline and intend to open-source the dataset, prompts, evaluations for the community to evaluate the models for manufacturing task guidance applications. 
We envision this work being part of larger efforts within the community to measure the generalization abilities of Large Multi-modal Models \cite{hupkes2023taxonomy}. 


\section{Related work}

Several large multimodal datasets relevant for multimodal process task guidance exist \cite{sener2022assembly101,damen2018scaling,wang2023holoassist,tang2019coin,alayrac2017learning,zhou2018towards}. Table~\ref{tab:dataset-comparison} shows comparison of dataset and their relevance for our manufacturing setting. The spec documents are missing in most of these datasets. Assembly101 \cite{sener2022assembly101} contains a pictorial representation of how the task could be performed. A natural language description is missing. In addition to this presence of multiple mounted camera view, synchronized over time makes it most relevant for our setting. In addition to the action labels, the dataset also contains `mistakes' in task execution which is important for measuring task guidance capabilities. However, the narrations of the task are missing. 

One of the intended applications of the work is contribute towards development of a benchmark to measure the Large model abilities in process task guidance scenarions. Benchmarks are intended to measure the performance of a system for a problem under study to compare systems subjected to similar conditions. 
\cite{hupkes2023taxonomy} develop a taxonomy for studying generalization in Large Language models. 
According to the taxonomy \cite{hupkes2023taxonomy} a strong motivation (Practical, cognitive, Intrinsic, Fairness \& inclusivity), type data shift (Covariate, Assumed, or both) and, type of generalization (structural, compositional, domain, robustness, language \&/or task) among others are necessary contributions for achieving generalization. This work fits the taxonomy by contributing an extended dataset that helps achieve a new practical use case, covariate shift and a novel domain. 

Each domain presents unique challenges. 
As a result of which several domain specific benchmarks for LLM evaluations have been released in recent times in order to track the progress. 
For instance, Legal \cite{hendrycks2021cuad,guha2024legalbench,fei2023lawbench,dai2023laiw}, Medical \cite{adams2024longhealth,dada2024clue,liu2024large,aali2024benchmark}, Finance \cite{wang2023fingpt,xie2024finben,hirano2024construction,xie2023pixiu,li2024alphafin,xu2024superclue} and, Education \cite{welbl2017crowdsourcing,rothe2021simple} among others. 
Benchmarks can even be limited in their abilities to measure progress and even leading to gamification \cite{schlangen2021targeting,choi2023llms,li2024crowdsourced}. 
They're nevertheless useful for researchers and developers studying specific domains and solving problems that are idiosyncratic to specific areas. 
While benchmarking practices have contributed to an uplift in performance of LLM models, recent research has highlighted that there is work to be done in ensuring that such benchmarks adequately represent the contexts into which LLMs are being deployed. This research extends the "everyday auditing" practices which have begun to emerge in which user communities identify AI-produced harms \cite{shen2021everyday,kuo2024wikibench}. 

\begin{wraptable}{r}{8.5cm}
\centering
\resizebox{0.55\columnwidth}{!}{
\begin{tabular}{cp{0.9cm}ccc}\\\toprule  
Dataset & Mistake tags & Narration & Actions & Camera\\\midrule
 \cite{damen2018scaling}   & No & No & Yes & Ego \\  \midrule
 \cite{tang2019coin}       & No & Yes & Yes & Varies \\  \midrule 
 \cite{zhou2018towards}    & No & Yes & Yes & Varies \\  \midrule
 \cite{wang2023holoassist}  & Partial & Yes & Yes & Ego \\  \midrule
 \cite{sener2022assembly101} & Yes & No & Yes & Mounted \\  \midrule
 Ours (Internal)            & Yes & Yes & Yes & Mounted \\  \bottomrule
\end{tabular}
}
\caption{Shows the comparison of a few selected datasets relevant for task guidance. Our dataset is the only one with a \textit{spec}. Mounted camera view is important since, the technicians cannot wear a camera in a precision manufacturing setting.}\label{tab:dataset-comparison}
\end{wraptable} 




\begin{algorithm}
\caption{Algorithm for QA-TOOLBOX}\label{alg:cap}
\begin{algorithmic}
\Require $Specs, Narrations, Instructions$
\State 1. Extract semantic entities from specs \& narrations 
\State 2. Anonymize Spec ($Anonymized_{spec}$) \& Narrations ($Anonymized_{narration}$) - Replace semantic entities with placeholders.
\State 3. Generate $Prompt_{narration}$ \& $Prompt_{spec}$. Use $instructions$, $Anonymized_{spec}$ and, $Anonymized_{narration}$
\State 4. Generate $Assembly101_{spec}$ \& $Assembly101_{narration}$ using LLM
\State 5. Generate $Questions$ dataset. 
\State 6. Use $Questions$, $Assembly101_{spec}$, $Assembly101_{narration}$, $Instructions$ to generate answers. 
\State 7. Use LLM-as-a-Judge to rate the responses using $Judge\_prompts$
\end{algorithmic}
\end{algorithm}

\section{Dataset Generation}
In this work we create QA-TOOLBOX for manufacturing. QA-TOOLBOX is a dataset built on top of \cite{sener2022assembly101}  to bridge the gaps in the earlier dataset (Table~\ref{tab:dataset-comparison}). Figure~\ref{fig:workflow} shows the overall workflow in creating the dataset \& automatic evaluation pipeline.
One of the major challenges for evaluating LLM viability for the manufacturing domain is a lack of a representative dataset validated in a real-world setting. In this work we bridge this gap by using real-world manufacturing data and have LLMs generate the missing information. 

A process task guidance dataset relevant for manufacturing task guidance is characterized by: (i) a task performer (technician) who performs the task, (ii) spec documents technicians follow to achieve a goal, (iii) task executions \&/or narrations where a technicians describes the actions in a given process.


\begin{figure}
    \centering
    \includegraphics[width=\linewidth]{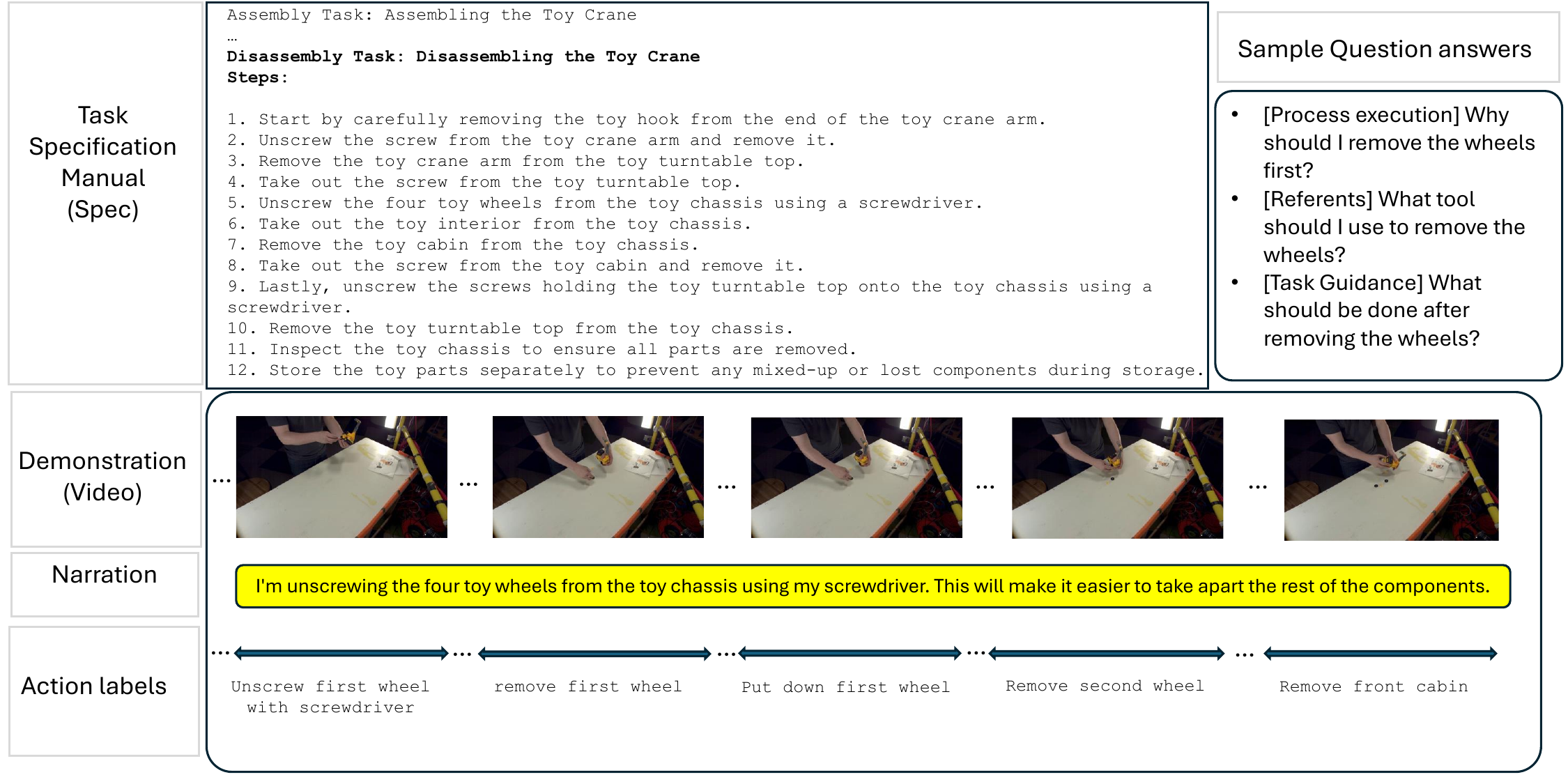}
    \caption{Figure shows an example from the QA-TOOLBOX. Demonstrations and action labels is accompanied by spec, narrations and question answers.  }
    \label{fig:example}
\end{figure}
\subsection{Dataset}
\paragraph{Assembly101} 
The choice of Assembly101 \cite{sener2022assembly101} was guided by our observations that the dataset matches most closely to our internal manufacturing dataset. 
Assembly101 is made up of multiple actors performing structured procedures captured from several camera viewpoints. 
Despite the structured procedure, the data consists of natural sequence variations due to the performer's skill level and various anomalies. The dataset is a large-scale video dataset consisting of annotations for action recognition (coarse \& fine grained), action anticipation, temporal action segmentation and mistake detection \cite{sener2022assembly101}. The dataset consists of people assembling and disassembling 15 different toys with 90 objects (tools \& toy parts).
and 24 interaction verbs comprising 1380 fine-grained action labels. 
However, this dataset doesn't contain textual instruction manuals for the assembly/disassembly sequence. The dataset also doesn't contain narrations of the task performer performing the actions. 

\paragraph{Manufacturing Dataset (Internal)} Our internal dataset collected from an advanced manufacturing site consists of spec documents detailing the manufacturing procedure, free form natural language narrations with the task performer describing the actions being performed during the process demonstration along with videos from mounted cameras. 
We collected data internally using a popular \textit{wizard-of-oz (WoZ} paradigm \cite{dahlback1993wizard,devault2014simsensei} where a remote-controlled agent interacted with the task performer during the data collection process. The technicians follow the spec to complete a process that involves steps around assembly, disassembly and cleaning. 
During the data collection process we observed that the technicians often forget to provide necessary information or even forget to narrate. A WoZ system helps nudge the technicians narrate and provide the information. The narrations collected are converted to text using Whisper speech to text \cite{radford2023robust}. 
We utilize 20 such recordings for prompting the LLMs to generate the augmentation dataset. 

\subsection{Augmenting missing data} 
Data augmentation has been done with LLMs to annotate, estimate quality, and generate new training samples \cite{ding2024data}. The quality of the generated content from LLMs has been shown to be comparable or even better than those generated by crowd-workers \cite{wu2023llms}. With the promise of such data augmentation approaches, we set out to extend the literature by using LLMs to fill the missing information in an existing dataset for manufacturing use cases. 

\subsubsection{Technicians narration data}
We collected a small sample ($n=20$) of interactions between technicians and a remotely controlled agent.
The narrations collected are intended to collect additional information typically not found in specs such as variants of referring expressions intentions for performing actions, and nuances of actions. 
We include technicians engaged in the manufacturing processes in our data collection .

\subsubsection{Participatory QA Dataset}  
We interviewed the entire cohort of task technicians within our partnering manufacturing facility and gathered questions they wanted to ask the agent in the course of their real-world work. 
(Table~\ref{tab:data_stats}) 
We then used their questions to seed augmentation. 
This participatory approach assures that our dataset is less speculative, and more promising for evaluating LLM performance in ways that humans actually need. 
This interviewing was conducted in a semi-structured protocol designed to assess trust, trustworthiness, and usability of building an AI assistant for manufacturing. 
In unstructured follow-ups, the interviewer asked the technicians to share all of the questions they would want to ask using voice-based UI in the course of their work. 

\subsubsection{Anonymization}
IP leakage in manufacturing facilities is important to be addressed. The exact protocols, objects, actions (duration), sequences of actions etc., are closely guarded. To ensure no such information are viewed by LLMs (locally deployed or cloud-based) we remove semantic entities and replace with placeholder, i.e., ([ACTION], [TOOL], [OBJECT], [MANNER], [TEMPORAL], [PURPOSE], [GOAL], [NOTE]). The LLMs are then prompted to generate narrations and spec similar in structure but for the Assembly101 dataset. 

\subsection{LLM Augmentations}

For augmentation, we use (from the internal dataset) (i) Specs that include description of process steps, (ii) Narrations similar to those of technicians describing actions, and finally, (iii) Questions. 

\subsubsection{Prompt generation}
\label{susubsec:prompt_gen}
We create prompts for the specs, questions and narrations with samples from the internal dataset and the assembly101 dataset annotations. The prompt templates are shown in Section~\ref{subsec:augmentation_prompts}. At the core of the prompt generation we use, $Prompt_{spec}$ (for creating specs), $Prompt_{narration\_creation}$ (for creating narrations), $Prompt_{q\_gen\_spec}$ and $Prompt_{q\_gen\_narration}$ (for generating questions) as the prompts to create new samples.
$Assembly101_{spec}$, $Assembly101_{narration}$ referring to the newly created specs and narrations for Assembly101 dataset. $\{Anonymized_{spec}\}$ and $\{Anonymized_{narrations}\}$ are the internal manufacturing spec with anonymized semantic entities. [$Inst\_pre$] is the instruction prefix appended in front of the prompt to assist LLMs in generating the samples. 

\begin{table}[]
    \centering
    \begin{tabular}{p{1.5cm}p{4.7cm}}
    
         \textbf{Category} & \textbf{Examples}\\ \hline         
          Process execution  & From the spec it is unclear how much the turntable top should be rotated or tilted, so could you specify how these motions should be performed for accurate assembly? \\
          Referents  & Where should the disassembled or assembled toy crane be stored? \\
          Task guidance & Do I inspect the crane after each step to ensure all parts were properly attached and tightened? \\
          Social     & Who developed this system? \\ \hline
    \end{tabular}
    \caption{Example questions for each of the categories}
    \label{tab:example_questions}
\end{table}

\begin{table}[]
    \centering
    \begin{tabular}{lcc}
    
         \textbf{Category} & \textbf{Assembly101} & \textbf{Internal} \\ \hline
          Referents  & 112,500 & 467 \\
          Procedure  & 64,188 & 335 \\         
          Task guidance & 31,178  & 290 \\
         \textbf{Total} & 200,841 & 1460 \\ \hline        
    \end{tabular}
    \caption{Categories across which the QA abilities are measured in a task guidance system.}
    \label{tab:data_stats}
\end{table}

\subsubsection{Spec, Narrations \& Questions}
We augment the Assembly101 dataset by creating specs and narrations using prompts (Sec~\ref{susubsec:prompt_gen}). 
We use an Instruction-tuned mixture of expert models and Mixtral-8x7b \cite{jiang2023mistral}.
We prompt the LLMs to generate narrations in a  style similar to the action labels in the Assembly101 dataset. 
In our qualitative analysis, we found the narrations and spec we generated had similar characteristics to those of our internal dataset. While no quantitative studies were conducted, an avenue for future work is to study the quality of narrations and spec documents generated. Since it has been shown that the quality of the generated augmented dataset can exceed that of human generated, we believe the same results holds true \cite{wu2023llms,ding2024data,li2024crowdsourced}. 
The spec and narrations had to be generated prior to generating the questions 
as the augmented specs and narrations are used to create the questions for our benchmark. 
We once again leverage Mixtral-8x7b model to generate questions (Refer Sec~\ref{subsec:augmentation_prompts} for prompt templates). 

\paragraph{Questions} We observed that the technicians often asked questions that belonged to four different categories based on the goal they're trying to accomplish and the importance of different inputs (spec, narration, instructions) and, modalities (text, images/video) for answering the questions. 
(i) \textbf{Process Executions}: Technician wants information about performing the process. They're are interested in elements of executions of the process such as tools required, ordering of steps, rationale for certain actions or usage of tools etc. 
This type of question requires spec, narration (if available) and common sense reasoning to answer.
(ii) \textbf{Referents}: Technician wants to disambiguate or to get information about referring expressions (typically action or noun phrases). These are the questions specifically about the objects, tools, locations or even the actions themselves. This type of questions requires spec and narrations to answer. Vision modality may help. 
(iii) \textbf{Task guidance}: Technician wants information about the next step or regarding the mistakes they committed w.r.t the action being performed. These are questions specific to performing tasks such as a sequence of steps, rationale for a given order and even what to do next. This type of input relies on spec and narrations especially in manufacturing domain. Commonsense reasoning may sometimes conflict with the recommended approach. Vision modality may help.  
We also identifies \textbf{Social} questions are not related to the task. Since, the answers to the social questions are not grounded in the spec document, we will not discuss the responses to these categories further. Table~\ref{tab:example_questions} shows examples.


%
\begin{figure*}
    \centering
    \includegraphics[width=.45\columnwidth, page=1, clip, trim=0 0 1cm 0]{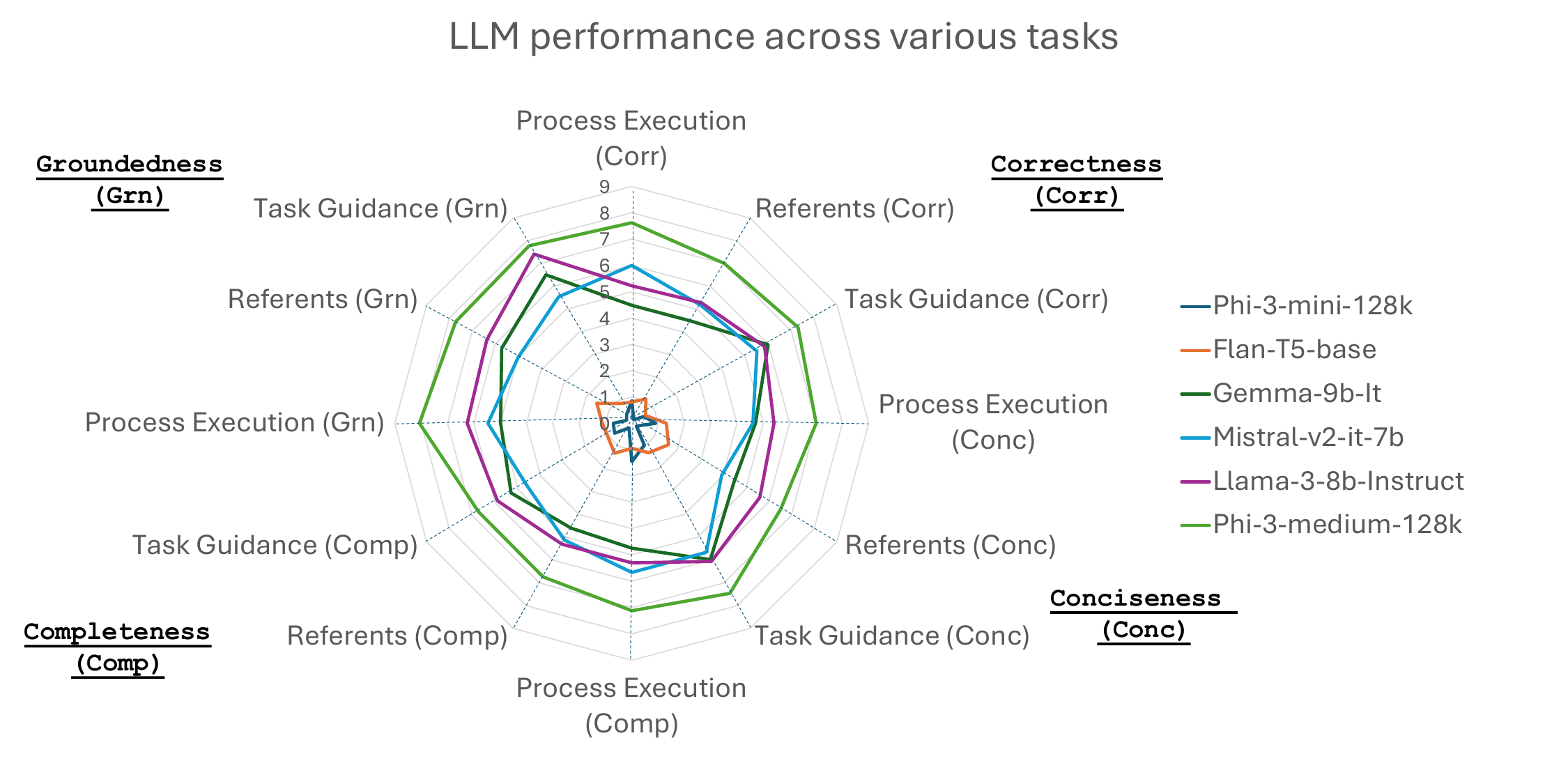}
    \includegraphics[width=0.45\columnwidth, page=1, clip, trim=2cm 0 2cm 0]{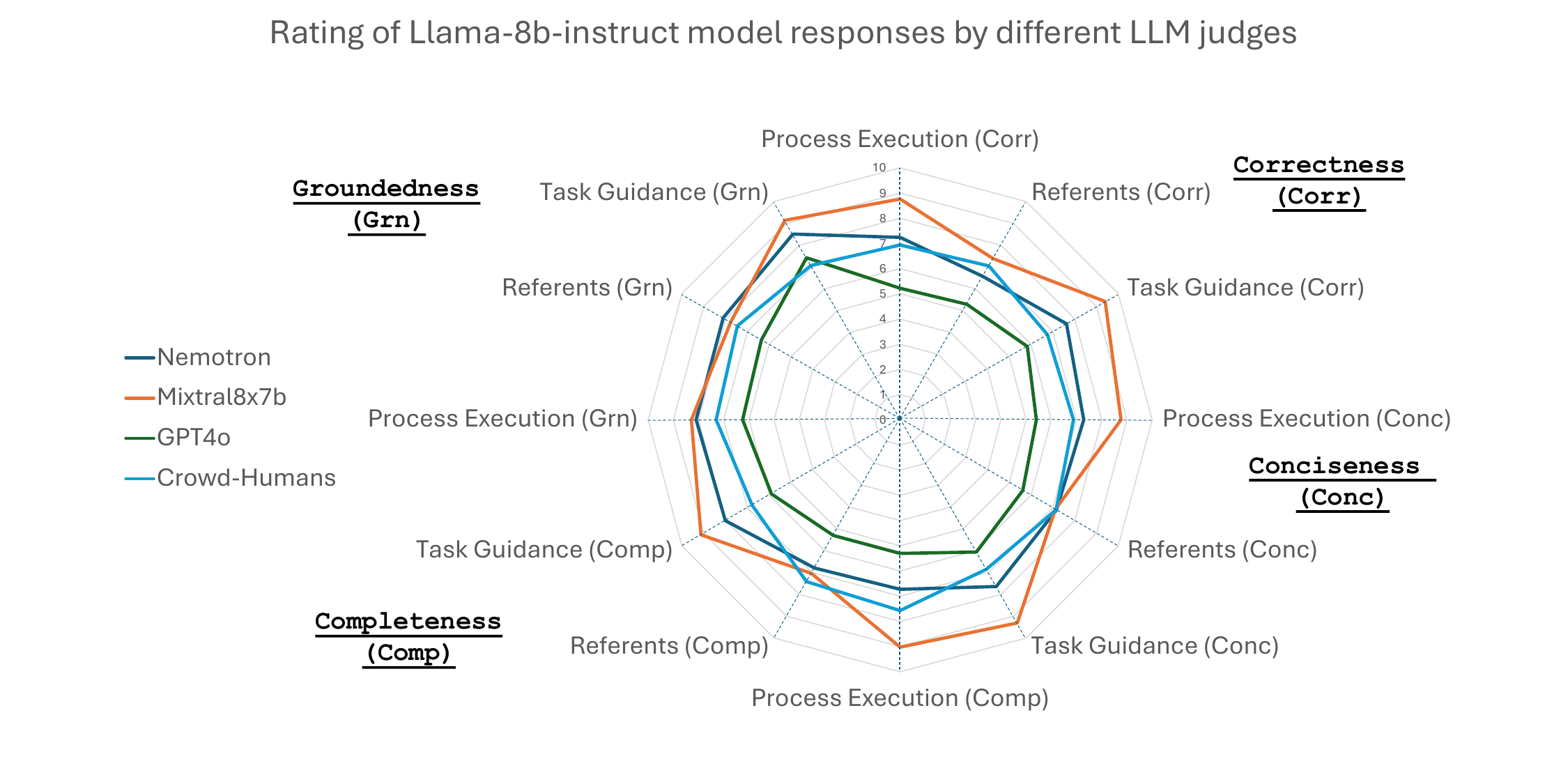}
    \caption{Radar charts of LLM-as-a-judge for the six models. i) Left chart shows  different models' performance on 3 categories of questions measured across Correctness, Conciseness, Completeness and Groundedness. Table~\ref{tab:model_comparison} shows the values captured in the charts.  ii) Right chart shows rating by different LLM-judge models for responses generated by LLama-3-8b-Instruct model. }
    \label{fig:results_model}
    \end{figure*}

\section{Baseline System}

We utilize the existing LLMs which have topped the LLM benchmarks in the recent past for baseline task guidance QA system. 
We confine our model selection to fewer than 15 billion parameters due the hardware limitations required to run larger models (in fp16 format)\footnote{Manufacturing organizations are concerned about IP leakage and costs and thus operating in a local non-cloud deployment with cost constraints is important}. 

\paragraph{Models}
We evaluate the LLMs that have achieved impressive zero shot performance in various benchmarks namely, Phi3 (mini-128k \& medium-128k) \cite{abdin2024phi}, Mistralv02-it \cite{jiang2023mistral}, Llama3-8b-Instruct \cite{llama3modelcard}, Gemma2-9b-Instruct \cite{gemma_2024} and, Flan-T5-base \cite{flant5} models to generate the answers. To answer the questions, the models are provided prompts consisting of instructions, spec and narrations. The models are not fine-tuned to the domain and used off-the-shelf.

\paragraph{Task Prompts} 
The task prompts are different from those used for generating the dataset. When generating the augmentation datasets, the prompts consists of ground  truth data (full Narrations, $Assembly101_{actions}$) data. However, when answering the questions at time `t', during the inference time the system only has access to the spec ($Assembly101_{spec}$) and any narrations that might be available from time `t-k' ($Assembly101_{narration}^{t-k}$). Since, the models can consume language-only input, we release the evaluations from unimodal scenario. We leave it to the future work to effectively use the vision modality to improve the results in the benchmark. 

The prompt used across the models is formatted as a chat interaction as is the standard with LLMs. Prompt = [Technician: $Assembly101_{narration}^{t-k}$, Technician: `\textit{You're a task assistant helping technicians on the factory floor to answer questions about a process}'. The spec for the process: $Assembly101_{spec}$. Now answer the question: $Question$]. 


\section{Evaluation}

Several viable approaches have been developed for evaluating QA systems in recent times \cite{kwiatkowski2019natural,bai2021more}.
Reference-based (Ground truth answers exist) and reference-free (Ground truth answers do not exist) metrics exist for evaluating the performance of the models for the QA tasks.
Evaluating the answers in a reference-based setting from the models is challenging since a question may have more than one correct answer. The answer must be adherent to the spec document overall but the answers can indeed vary either in their surface form or even content-wise but still be adherent to the spec. 
Classical reference-free metrics rely on statistical patterns that exist in the questions, answers and document resulting in unreliable characterization of model capabilities \cite{deutsch2022limitations}. 
Recently, LLM-as-a-judge \cite{zheng2024judging} has gained popularity. LLMs as judge have been shown to achieve high correlation or even outperforming the crowd human ratings \cite{zheng2024judging,wu2023llms}.  
We utilize LLM-as-a-judge paradigm to evaluate the models in TOOLBOX. However, the question remains as to which LLM acts as the best judge? 
We experiment with three different LLMs and evaluate the ratings assigned by each. Finally, to validate the LLM-as-a-judge paradigm, we also collect crowd-sourced rating of the responses generated by one of our LLMs in the baseline stage (llama3-8b-instruct).
We also conduct a study to validate the effectiveness of LLM-as-a-judge by comparing the scores generated by the LLMs and crowd-workers with experts. 

We evaluate the performance of the models on conciseness, correctness, completeness and groundedness of the responses generated. 
For manufacturing the conciseness and groundedness of the responses are important along with correctness and completeness. The answers needs to be short since the technicians are time-bound and often not interested in lengthy responses. The responses needs to be grounded in the spec/narrations since the responses must not deviate from the spec documents.

\section{Results}

We evaluate the answers using LLM-as-a-judge (Prompts can be found in Section~\ref{subsec:judge_prompts}). More specifically we answer the following questions, 

\textbf{R1: How do different LLMs as backbone for the judge compare?}
We use LLMs with $>$ 45 billion parameters as judges. We utilize Mixtral8x7b \cite{jiang2023mistral}, Nemotron-4-340B-Instruct \cite{adler2024nemotron} and GPT4o \cite{OpenAI_GPT4O} as judges.
For the same prompts(instruction, question \& answer) we observed that different LLMs generated different scores (Fig~\ref{fig:results_model}).
We also request for the explanations from the models for the generated scores. 
We find that \textbf{ while the explanations/feedback generated are consistent across Nemotron and GPT4o, GPT4o was more strict when rating the responses of the models.}
Mixtral was the most liberal in scoring the responses. The exact reason for difference in scoring is left for the future work. The choice for the best LLM for judge was driven by expert preferences (R2). 

\textbf{R2: How do the ratings generated by LLM as a judge compare with those generated by the crowd-workers? }
While different validation experiments could be conducted (e.g., inter rater agreement between expert ratings and LLM as judge scores, correlation b/w crowd-workers and LLM as a judge scores etc.), one important question is \textit{Could LLM-as-a-judge be utilized for evaluation in a reference-free setting instead of human (crowd-workers) ratings while still being agreeable to experts?} To answer this question, we request domain experts (2 Socio-technical \& 2 Computer Science researchers) to rate the LLM-as-a-judge and crowd-worker scores in a double blind setting. Specifically, given a "spec, question, answer and ratings" the experts rated if they agree to disagree with the ratings while being unaware of source of ratings. The experts were also request to pay attention to the scale. 
To be consistent, we utilize the ratings generted for Llama3 model responses. 
We also observe that the experts significantly preferred the ratings from the GPT4o \cite{OpenAI_GPT4O} as a judge compared to crowd-workers (Student's t-test, n=150, p=0.04). The experts were also found to agree with each other's ratings (Cohen's kappa = 0.61, N=50 indicating high agreements).  
Hence, we can infer that \textbf{LLM-as-a-Judge \cite{zheng2024judging} could be useful as an automated reference free metric for evaluating the model responses in TOOLBOX}.  

\textbf{R3: Which models perform the best? }
We observe that the responses generated by Phi-3-medium-128k (14 billion parameters) performed the best across the question categories. This is consistent with the notion that higher parameter count results in better performance. We observe that the performance of this model is consistently better across the categories and type of questions.  Interestingly, mini version of the same model was the worst performing in our comparison. We did not analyze the causes for this difference in performance. Out of the models fewer than 10 billion parameters Llama3-8b-Instruct was the best performing compared to Gemma-2 and Mistral-7b models. (Figure~\ref{fig:results_model})

The latency of the response generated increases as we move to deploy models with higher parameter count. In addition constraints of H/W and cost limitations also need to be taken into account. Hence, it is also important to be mindful of the latency and take informed decisions about these constraints. Hence, the best possible model to deploy is also constrained by hardware options available for the concerned. 

\section{Discussion \& Future work}

In this work we propose a novel approach to building and using benchmark datasets designed to advance pluralistic and socio-technical values in computing research (Refer Sec~\ref{sec:defs} for definitions). 
QA-TOOLBOX presents several promising direction for applied and fundamental research. 
The manufacturing task guidance is a complex problem requiring the models to generate recommendations grounded in spec documents. The sequence of actions to be performed are rigid, however allowing certain flexibility in certain places. The benchmark allows comparing open-sourced LLMs and proprietary models for manufacturing. The approach itself is scalable and applicable to several manufacturing processes. 


Furthermore, current multi-modal LLMs \cite{liu2023improvedllava,zhu2023languagebind,liu2023llava,lin2023video,laurenccon2024obelics} can ingest a few frames of visual context limiting their capability to understand long-term visual context and reasoning. This long-term visual context and reasoning is critical for an efficient task guidance system that needs to provide timely feedback for the natural/common step execution variations such as repeated steps or uncompleted steps in the distant past resulting from unforeseen scenarios.



For the future versions of the benchmark, we intend to release updated tasks (summarization, continual learning, multimodal tasks) which are relevant for manufacturing domain. RAG-based \cite{lewis2020retrieval} approaches are central to developing such systems either to prevent hallucinations or reducing the context when the spec documents reach several thousand pages in length.

\bibliographystyle{plainnat}
\bibliography{custom}

\appendix

\section{Appendix}
\label{sec:appendix}

\begin{table*}
    \centering
    \resizebox{\textwidth}{!}{%
    \begin{tabular}{l|ccc|ccc|ccc|ccc}
        \hline
        & \multicolumn{3}{c|}{Correctness} & \multicolumn{3}{c|}{Conciseness} & \multicolumn{3}{c|}{Completeness} & \multicolumn{3}{c}{Groundedness} \\
        GPT4o-as-Judge & PE (Corr) & Ref (Corr) & TG (Corr) & PE (Conc) & Ref (Conc) & TG (Conc) & PE (Comp) & Ref (Comp) & TG (Comp) & PE (Grn) & Ref (Grn) & TG (Grn) \\
        \hline
        Phi-3-mini-128k & 0.87 & 0.15 & 0.53 & 0.92 & 0.20 & 0.96 & 1.45 & 0.20 & 0.75 & 0.70 & 0.23 & 0.39 \\
        Flan-T5-base & 0.81 & 1.07 & 0.60 & 1.31 & 1.62 & 1.29 & 0.95 & 1.31 & 1.05 & 1.12 & 1.53 & 0.87 \\
        Gemma-9b-It & 4.48 & 4.49 & 5.98 & 4.70 & 4.48 & 5.97 & 4.75 & 4.60 & 5.30 & 4.98 & 5.70 & 6.52 \\
        Mistral-v2-it-7b & 6.00 & 5.20 & 5.49 & 4.60 & 3.94 & 5.66 & 5.66 & 5.12 & 4.63 & 5.46 & 4.99 & 5.56 \\
        Llama-3-8b-Instruct & 5.22 & 5.29 & 5.82 & 5.41 & 5.62 & 6.06 & 5.30 & 5.30 & 5.89 & 6.25 & 6.35 & 7.43 \\
        \textbf{Phi-3-medium-128k} & \textbf{7.62} & \textbf{7.02} & \textbf{7.30} & \textbf{7.01} & \textbf{6.53} & \textbf{7.45} & \textbf{7.12} & \textbf{6.74} & \textbf{6.72} & \textbf{8.07} & \textbf{7.73} & \textbf{7.79} \\
        \hline
    \end{tabular}%
    }
    \caption{Comparison of various models across different metrics by LLM (GPT-4o) as a judge on a scale of 0 - 10. }
    \label{tab:model_comparison}
\end{table*}

\subsection{Augmentation prompts}
\label{subsec:augmentation_prompts}
$Prompt_{spec}$ = [$Inst\_pre$] Here is a sample spec document $\{Anonymized_{spec}\}$. Create a similar spec document where the following $Assembly101_{actions}$ were observed. (1)

$Prompt_{narration\_creation}$ = [$Inst\_pre$] Here is a sample spec document $\{Anonymized_{spec}\}$. A technician following the spec generated the following narrations $\{Anonymized_{narrations}\}$. Create a similar narrations for $Assembly101_{spec}$ . (2)

$Prompt_{q\_gen\_spec}$ = [$Inst\_pre$] Here is a sample spec document $\{Anonymized_{spec}\}$. A technician asked the following questions $\{Anonymized_{questions}\}$. Create similar questions for $Assembly101_{spec}$. (3)

$Prompt_{q\_gen\_narration}$ = [$Inst\_pre$] Here is a sample narrations from a technician $\{Anonymized_{narrations}\}$. A technician is interested in asking the following questions $\{Anonymized_{questions}\}$. Create similar questions for $Assembly101_{narration}$. (4)

[$Inst\_pre$] = \{`Please Generate n samples', `Be creative', `You can use some generally used synonyms', `Be adventurous'\}. (5)

\begin{figure}
    \centering
    \includegraphics[width=\columnwidth]{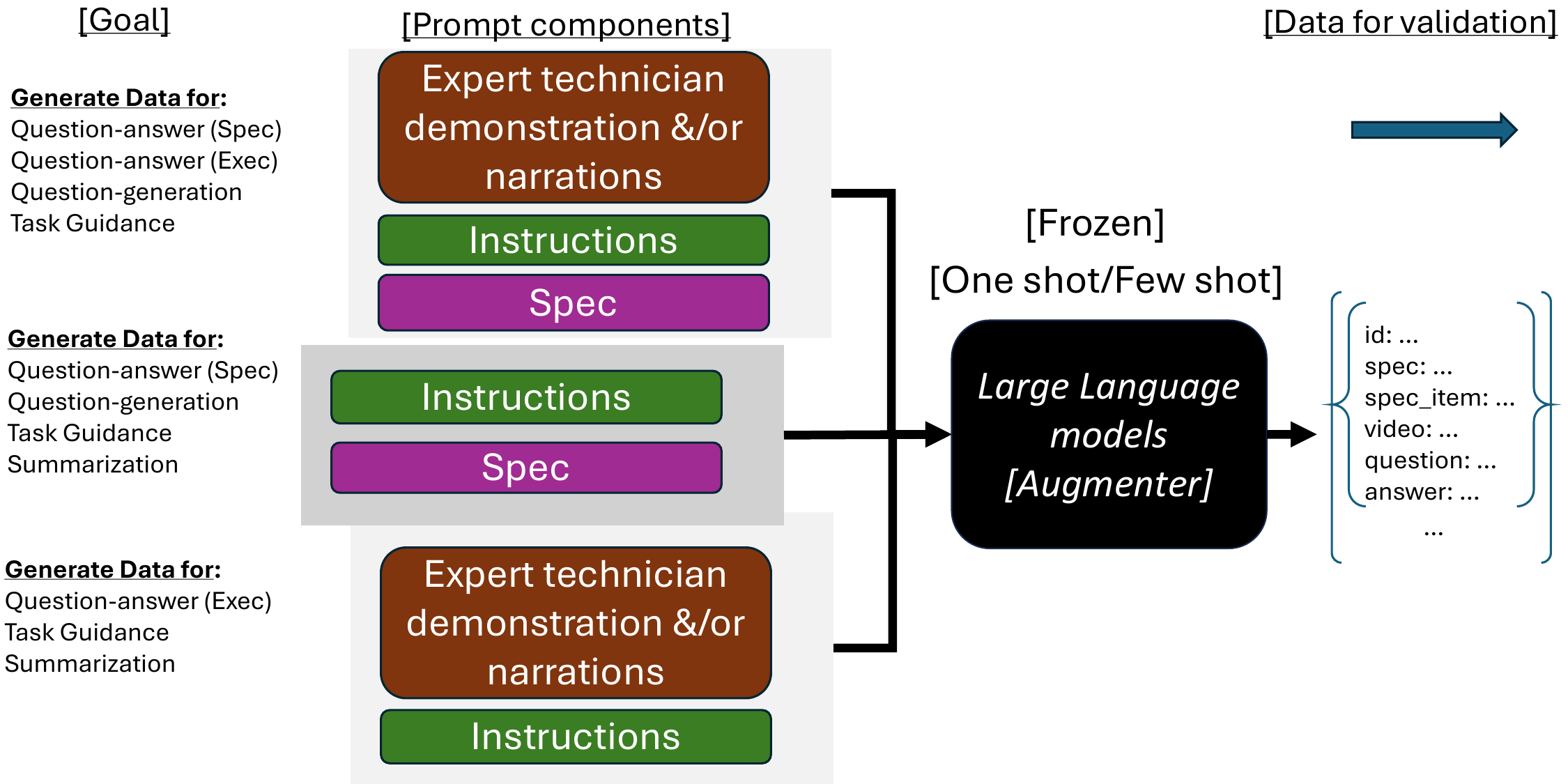}
    \caption{Shows the three different augmentation approaches and format of the data}
    \label{fig:enter-label}
\end{figure}

\subsection{Judge Prompts}
\label{subsec:judge_prompts}
\textbf{completeness\_prompt} = """
        You will be given a question and answer couple.
        Your task is to provide a 'total rating' on completeness and score how well the answer answers the user concerns expressed in the user\_question.
        Give your answer as a float on a scale of 0 to 10, where 0 means that the response is not complete at all, and 10 means that the answer is complete, concise and addresses the question. Deduct points for not being complete in the response. Use spec as a reference.  

        Provide your feedback as follows:

        Feedback:::
        Total rating: (your rating, as a float between 0 and 10)

        Now here are the question and answer.
        Spec: {spec}
        Question: {question}
        Answer: {answer}

        Feedback:::
        Total rating: """
\\ \noindent      
        \textbf{conciseness\_prompt} = """
        You will be given a question and answer couple.
        Your task is to provide a 'total rating' scoring how well the answer answers the user concerns expressed in the user\_question.
        Give your answer as a float on a scale of 0 to 10, where 0 means that the response is not concise at all, and 10 means that the answer is complete, concise and addresses the question. Deduct points for not being to the point or generating verbose responses. Answers should always be to the point. Use spec as reference. 

        Provide your feedback as follows:

        Feedback:::
        Total rating: (your rating, as a float between 0 and 10)

        Now here are the question and answer.
        Spec: {spec}
        Question: {question}
        Answer: {answer}

        Feedback:::
        Total rating: """
\\ \noindent
        \textbf{correctness\_prompt} = """
        You will be given a question and answer couple.
        Your task is to provide a 'total rating' scoring how well the answer answers the user concerns expressed in the user\_question.
        Give your answer as a float on a scale of 0 to 10, where 0 means that the response is not correct at all, and 10 means that the answer is complete, correct and addresses the question. Deduct points for not being accurate or including any irrelevant phrases or sentences. Use spec as a reference. 

        Provide your feedback as follows:

        Feedback:::
        Total rating: (your rating, as a float between 0 and 10)

        Now here are the question and answer.
        Spec: {spec}
        Question: {question}
        Answer: {answer}

        Feedback:::
        Total rating: """
\\ \noindent
        \textbf{Groundedness\_prompt} = """
        You will be given a question and answer couple.
        Your task is to provide a 'total rating' scoring how well the answer answers the user concerns expressed in the user\_question based on novelty. The answer should not be very novel. 
        Give your answer as a float on a scale of 0 to 10, where 0 means that the answer is very novel and is unrelated to the question, or spec. 10 means that the answer is not novel and is totally grounded in the spec. Deduct points for being very novel or including any irrelevant phrases or sentences. Use spec as a reference. 

        Provide your feedback as follows:

        Feedback:::
        Total rating: (your rating, as a float between 0 and 10)

        Now here are the question and answer.
        Spec: {spec}
        Question: {question}
        Answer: {answer}

        Feedback:::
        Total rating: """

\section{Some definitions}
\label{sec:defs}
"Sociotechnical" is a term that helps researchers recognize that technical systems are deployed into social systems, and so technical systems will be both constrained and enabled by social systems such as organizations, hierarchies, and workflows \cite{leonardi2012materiality}. Recent calls for "pluralism" in benchmarking describe that such practices need to better encompass diverse human values and needs \cite{sorensen2024roadmap}. Both of these approaches apply to the internal dataset regarding the manufacturing task, to avoid risks of IP (Intellectual Property) leakage.

\end{document}